\documentclass[5p,times]{elsarticle}

\usepackage{ecrc}
\usepackage{booktabs}
\usepackage{caption}

\journalname{Manufacturing Letters}

\runauth{S. Li and C. Shao}

\usepackage{amssymb}
\usepackage{amsthm}
\usepackage{multirow} 
\usepackage{tabularx}

\usepackage[figuresright]{rotating}

\usepackage[bookmarks=false]{hyperref}
    \hypersetup{colorlinks,
      linkcolor=blue,
      citecolor=blue,
      urlcolor=blue}

\usepackage{soul}

\title{Multi-Modal Data Fusion for Moisture Content Prediction in Apple Drying}

\begin{document}

\begin{frontmatter}



\title{Multi-Modal Data Fusion for Moisture Content Prediction in Apple Drying}


\author[a]{Shichen Li} 
\author[b,a]{Chenhui Shao\corref{cor1}}
\ead{chshao@umich.edu}

\address[a]{Department of Mechanical Science and Engineering, University of Illinois at Urbana-Champaign, Urbana, IL 61801, USA}
\address[b]{Department of Mechanical Engineering, University of Michigan, Ann Arbor, MI 48109, USA}

\begin{abstract}
Fruit drying is widely used in food manufacturing to reduce product moisture, ensure product safety, and extend product shelf life. Accurately predicting final moisture content (MC) is critically needed for quality control of drying processes. State-of-the-art methods can build deterministic relationships between process parameters and MC, but cannot adequately account for inherent process variabilities that are ubiquitous in fruit drying. To address this gap, this paper presents a novel multi-modal data fusion framework to effectively fuse two modalities of data: tabular data (process parameters) and high-dimensional image data (images of dried apple slices) to enable accurate MC prediction. The proposed modeling architecture permits flexible adjustment of information portion from tabular and image data modalities. Experimental validation shows that the multi-modal approach improves predictive accuracy substantially compared to state-of-the-art methods. The proposed method reduces root-mean-squared errors by 19.3\%, 24.2\%, and 15.2\% over tabular-only, image-only, and standard tabular-image fusion models, respectively. Furthermore, it is demonstrated that our method is robust in varied tabular-image ratios and capable of effectively capturing inherent small-scale process variabilities. The proposed framework is extensible to a variety of other drying technologies.
\end{abstract}

\cortext[cor1]{Corresponding author.}

\end{frontmatter}



\section{Introduction}

Fruit drying is an important process in food manufacturing. It plays a key role in long-term preservation while facilitating storage and reducing transportation costs by removing moisture and maintaining the integrity of fruits~\cite{omolola2017quality}. However, the fruit drying process is highly intricate, involving numerous input variables and diverse output objectives~\cite{castro2018mathematical}. Key input variables include environmental drying process parameters and natural variations among fruit samples~\cite{defraeye2017convective, trirattanapikul2016influence}; and quality control objectives include moisture levels, fruit tastes, water activity levels, etc.~\cite{defraeye2017convective, defraeye2018insights, raponi2017monitoring}. Given the challenges of collecting extensive drying data due to time and equipment constraints~\cite{chua2003low}, accurately modeling these complex input-output relationships with limited datasets is essential for meeting quality control goals~\cite{khan2022application}.

Various fruit drying studies have investigated different input-output relationships, examining how drying conditions and sample characteristics influence the physical drying processes and final product quality. For example, prior research has studied the effects of temperature on rehydration kinetics in cherry drying~\cite{turkmen2020effects}, the influence of air velocity on energy consumption in blueberry drying~\cite{yu2017effects}, and the impact of banana thickness on drying efficiency~\cite{kumar2019thin}. Among these studied relationships, using inputs, which are primarily process parameters, to predict final moisture content (MC) is a common objective, as it is crucial for ensuring product quality and improving efficiency of the fruit drying process~\cite{bourdoux2016performance, raponi2017monitoring}. Studies have shown that higher temperatures or air velocity reduce final MC, while longer drying times significantly lower MC for near-fresh fruits but have minimal impact once a certain dryness level is reached~\cite{defraeye2017convective}. Additionally, post-drying sample characteristics such as color, water activity, and texture can serve as indicators of MC~\cite{mathlouthi2001water, martynenko2014texture}. 

Accurately predicting final MC in fruit drying remains very challenging due to limitations in data diversity and a lack of modeling approaches capable of fully leveraging distinct data types (modalities). Most existing studies model MC based on controlled environmental process conditions (e.g., temperature, air velocity, drying time)~\cite{vega2012effect, santacatalina2014ultrasonically}, but overlook inherent sample-specific variations such as slight differences in color and texture among individual fruit samples. Such sample-specific variations are inevitable in industrial drying processes, where samples vary in random ranges of thickness, weight, diameter, and initial MC, introducing additional variabilities that are difficult and/or costly to measure a priori~\cite{li2025uncertainty}. Subsequently, the lack of capturing such variabilities leads to unsatisfactory modeling accuracy, which further limits the effectiveness of quality control actions. While some studies focused on modeling the impact of sample-specific characteristics on MC under consistent drying conditions, these models assume those sample characteristics are controllable and are less effective when applied to diverse drying environments~\cite{bora2018image}. A recent study extracts two simplified sample-specific features (average RGB color and area) from image data and then combining these features with environmental process parameters for MC prediction~\cite{keramat2021real}. Nevertheless, this research condensed high-dimensional image data to limited information, which may miss essential information contained in images. Additionally, the used machine learning models treated all variables equally, which potentially dilute the influence of critical environmental factors and limit the predictive capabilities.

Multi-modal data fusion is a method for leveraging multiple data formats from different sources, enabling their integration without pre-converting them into a single modality to enhance prediction capabilities~\cite{baltruvsaitis2018multimodal}. Recent advancements in multi-modal learning have demonstrated the significant potential of improving predictive accuracy and adaptability across various domains. For instance, in manufacturing, multi-sensor fusion has been utilized for online monitoring and quality prediction~\cite{wu2022end, meng2024meta, meng2023explainable, petrich2021multi, gaikwad2022multi}. In robotics, the integration of RGB and depth images has played a crucial role in facilitating robust decision-making, particularly in real-world tasks such as robotic grasping and autonomous delivery~\cite{cadena2016multi, sihite2024dynamic}.

However, multi-modal data fusion has been underexplored in drying research. One of the key challenges in applying multi-modal data fusion to fruit drying is the substantial disparity between different data modalities. This disparity makes it difficult for a single model or modeling architecture to process two or more fundamentally different data types without pre-transformation. For example, process parameters are structured, low-dimensional, and tabular numerical data representing fixed, manually controllable conditions such as temperature, air velocity and drying time. In contrast, sample-specific characteristics exhibit inherent variability, making high-dimensional images of drying samples a more informative data source. These images capture pixel-wise color and area information, which more accurately reflect key physical attributes. This richer information source offers a substantial potential for improving MC prediction accuracy. Therefore, it is essential to improve the model architecture to effectively integrates heterogeneous data modalities, ensuring that both process parameters and sample-specific characteristics contribute optimally to MC prediction.

\begin{figure}[h]
    \centering
    \includegraphics[width=1\columnwidth]{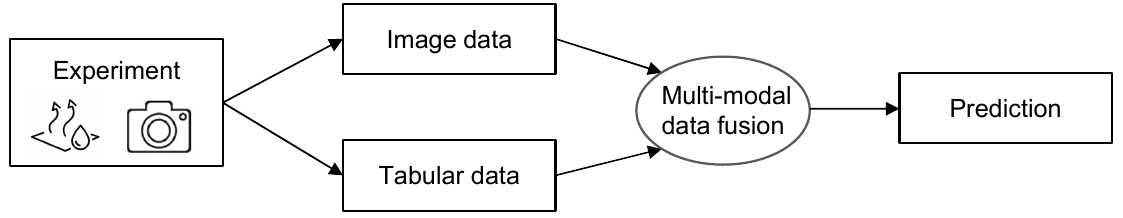}
    \caption{Schematic of the multi-modal data fusion framework for MC prediction.}
    \label{f1_framework}
\end{figure}

In this paper, we develop a novel multi-modal data fusion framework for predicting MC in apple drying. The overall framework is illustrated by Figure~\ref{f1_framework}. This research utilizes an apple drying dataset from experiments conducted on a remodeled air convective dryer. We use both tabular and image data to account for drying conditions and inherent variations in sample characteristics. Our method processes each data modality in parallel through models including fully connected (FC) neural network (NN) layers, segment-anything-model (SAM), and ResNet-18, and preserves high-dimensional image features. The proposed modeling architecture fully utilizes both tabular and image data by portion-adjustable concatenating without converting them into a single data type. Experimental studies demonstrate the effectiveness of the proposed framework--it reduces root-mean-squared errors (RMSEs) by 19.3\%, 24.2\%, and 15.2\% over tabular-only, image-only, and standard tabular-image fusion models, respectively. It is also shown that the proposed parallel processing approach not only captures sample-specific nuances that are often missed in single-modality models but also ensures effective information fusion with adjustable portion of diverse data configurations. The results highlight the effectiveness of this multi-modal fusion strategy for enhancing predictive accuracy in food drying applications.

Our work presents three key contributions that offer advantages over conventional approaches. First, we propose a novel multi-modal fusion framework that integrates structured process parameters with high-dimensional image data, significantly enhancing MC prediction accuracy in apple drying while providing new architectural insights applicable to other domains. It should be noted that most existing sensor fusion methods used in manufacturing struggle to effectively handle diverse modalities such as tabular and image data.
Second, our approach demonstrates high data efficiency by employing a novel data-splitting strategy that ensures non-overlapping training and evaluation sets, effectively managing small, unbalanced industrial datasets. This strategy improves the practical applicability and generalizability of our method in real-world drying scenarios. Third, we validate the necessity of incorporating high-dimensional image data by comparing our multi-modal fusion approach with standard data fusion methods, demonstrating its ability to capture complex and uncertain sample variabilities.

The remainder of this paper is organized as follows. Section 2 details the experimental design and data collection process. Section 3 describes the methodology, including the proposed multi-modal fusion framework, baseline models, and ablation study. Section 4 presents the results of applying this methodology to a case study on predicting MC in apple drying. Section 5 discusses the insights gained from the results in detail. Finally, Section 6 concludes the paper and suggests directions for future research.

\section{Design of experiments and data collection}

\subsection{Experimental design}

Fuji apples are used in this research. Before drying, each apple is cored using a 2.5 cm-diameter core remover and sliced transversely with an electric slicer to achieve a millimeter level thickness. The slices naturally vary in initial thickness, diameter, and weight, which aligns with standard industrial practices. The initial wet-based MC of the fresh apples is approximately 85\%.

\begin{figure}[h]
    \centering
    \includegraphics[width=0.83\columnwidth]{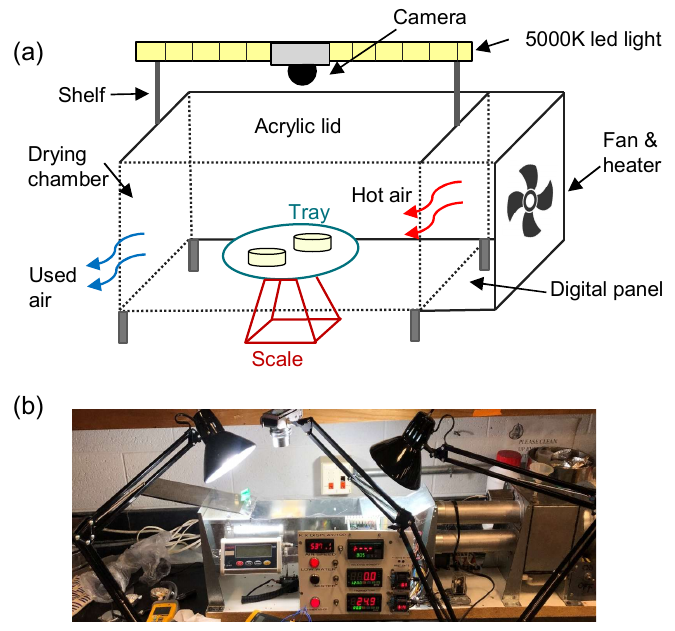}
    \caption{Experimental setup: (a) schematic model; and (b) photograph.}
    \label{f2_experiment1}
\end{figure}

The drying process employs a hot air convective dryer. Hot air convective drying is the most widely used in food drying~\cite{adeyeye2022food}. It removes moisture from the apple slices by transferring heat, mass, and airflow through the hot air stream~\cite{beigi2016hot}. We remodel the dryer to collect multi-modal data, with the details shown in Figure~\ref{f2_experiment1}. Figure~\ref{f2_experiment1}(a) provides a schematic of the dryer setup. The acrylic lid replaces the original lid; and LED lighting and a camera are mounted overhead the lid for in-situ image capturing. The air heater and fan generate a controlled hot air stream that flows over the apple slices placed on a tray within the drying chamber. A scale below the tray measures weight changes in real time, allowing for precise tracking of weight loss. Figure~\ref{f2_experiment1}(b) shows a photograph of the actual experimental setup. Each drying process includes either one or two apple slices, enabling us to explore small-scale variability in drying characteristics.

Drying experiments are conducted using a combination of three temperature levels (60$^{\circ}$C, 70$^{\circ}$C, 80$^{\circ}$C) and two air velocities (1.5 m/s, 2.5 m/s). To capture a broad range of final MC outcomes, we conduct experiments for each temperature-air velocity combination until the sample reached final MC levels clustered around 10\% and 20\%, respectively. The approximate final MC is determined by monitoring the sample weight using the scale, and the corresponding drying time, recorded as a process variable, varied from 70 to 250 minutes. For each combination of temperature, air velocity, and MC level, we conduct two repetitions of experiments with a single slice and three repetitions with two slices, yielding a total of 84 data samples. Additionally, drying samples’ characteristics, such as thickness and weight, are varying in ranges, which potentially lead to process variations. We record the initial and final weights of each apple slice to calculate moisture reduction during drying. We assume a consistent initial MC across fresh apples. Three slices are cut from each apple, and subsequently dried at 200°C until their weights stabilize. Then, the average MC is calculated based on weight reduction on these three slices. This average MC value is applied as the initial MC to all slices from the same apple.

\subsection{Dataset}

The collected dataset consists of both tabular and image data. Table~\ref{t1} outlines the tabular inputs, including temperature, air velocity, and drying time, with their ranges derived from the experimental design. In addition to the tabular data, we capture in-situ images of the apple slices throughout each drying process, and use the final image of each slice as the initial image data that serves as an input for the multi-modal data fusion framework. These images contain high-dimensional color and shape information that potentially reflects each sample's individual characteristics such as uncertain thickness, weight, diameter values which can contribute to the determination of the final MC. Moreover, this image data provides insights into subtle differences in drying outcomes that may not be evident from tabular data alone, which can potentially enhance the accuracy of final MC prediction.

\begin{table}[h]
\centering
\captionsetup{singlelinecheck=false, aboveskip=1pt}
\caption{Tabular data in the drying experiments.}
\vspace{6pt}
\footnotesize
\begin{tabular*}{\hsize}{@{\extracolsep{\fill}}p{3.5cm} p{3cm} p{3cm}@{}}
\toprule
Variables & Unit & Ranges \\
\colrule
Temperature & $^{\circ}$C & 60, 70, 80 \\
Air velocity & m/s & 1.5, 2.5 \\
Drying time & mins & 70--250 \\
\botrule
\end{tabular*}
\label{t1}
\end{table}

\begin{figure}[h]
    \centering
    \includegraphics[width=1\columnwidth]{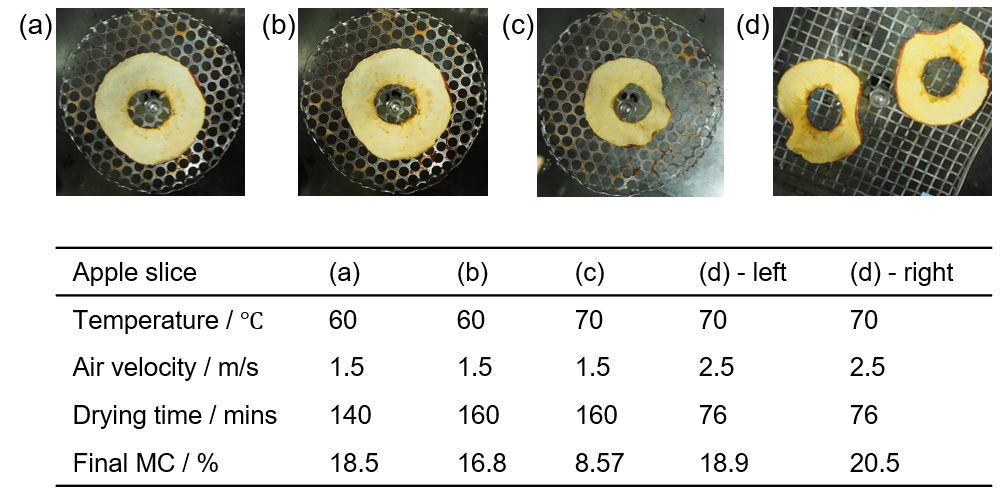}
    \caption{Examples of apple drying images under different drying conditions with different final MC values.}
    \label{f3_experiment2}
\end{figure}

Figure~\ref{f3_experiment2} presents examples of the captured images, with each panel showing a distinct drying condition and corresponding final MC. Comparing Figure~\ref{f3_experiment2}(a)--(d), we observe that slices subjected to different temperature ( Figure~\ref{f3_experiment2}(a) and Figure~\ref{f3_experiment2}(c)), air velocity ( Figure~\ref{f3_experiment2}(c) and Figure~\ref{f3_experiment2}(d)) and drying time ( Figure~\ref{f3_experiment2}(a) and  Figure~\ref{f3_experiment2}(b)) settings display noticeable variations in area and coloration, highlighting the influence of tabular data (i.e., process parameters) on the physical properties of the slices. As these samples have varying final MC values, their visual differences serve as effective indicators. In Figure~\ref{f3_experiment2}(d), slight differences in color and size are observed under the two slices dried simultaneously under identical drying conditions (70$^{\circ}$C, 2.5 m/s, 76 minutes). This illustrates how image data captures extra variability outside tabular data in drying effects.

The objective is to predict the final MC of apple slices after drying. We determine the ground truth of the final MC value of each apple slice based on its initial MC, initial weight, and final weight following Equation (\ref{eq:MC}).
\begin{equation} \label{eq:MC}
\mathrm{MC}_a = \frac{w_a - w_0 \times (1 - \mathrm{MC}_0)}{w_a},
\end{equation}
where $w_{0}$ and $w_{a}$ are initial and final weight of the apple slice, MC$_{0}$ and MC$_{a}$ are the initial and final MC of the apple slice. 

\section{Methodology}

This section presents multi-modal data fusion methodology to predict the final MC of apple slices in drying. Specifically, Section 3.1 details the architecture of the proposed multi-modal fusion framework; Section 3.2 describes the baseline models; and Section 3.3 explains the ablation study to evaluate the contributions of each data modality.

\subsection{Multi-modal fusion framework}

Figure~\ref{f4_model1} illustrates the proposed multi-modal fusion framework. Our model is designed to process tabular and high-dimensional image data in parallel, which preserves the unique characteristics of each data modality and allows for more effective feature extraction. The following provides a detailed explanation of how each data modality is processed and integrated within this network architecture:

\begin{figure*}[h]
    \centering
    \includegraphics[width=1.65\columnwidth]{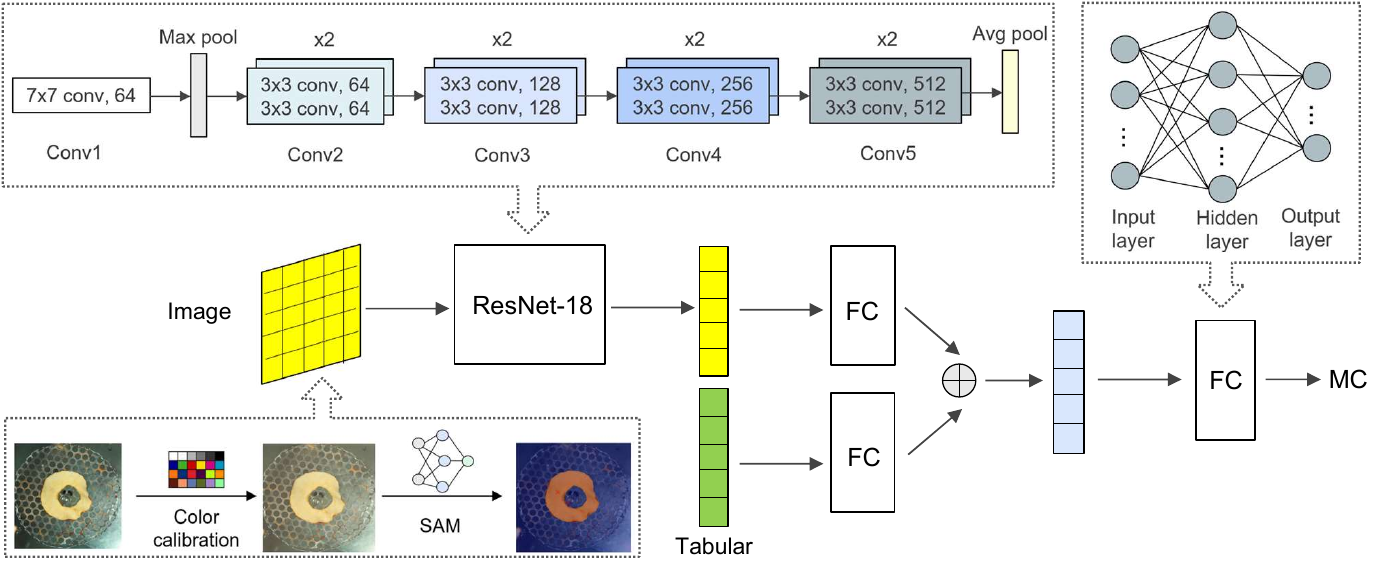}
    \caption{Multi-modal fusion framework.}
    \label{f4_model1}
\end{figure*}

\emph{Image data segmentation with SAM}: To prepare the image data, we perform color calibration and segmentation on each image for each apple slice. Color calibration adjusts the images from 5000K to the 6500K for consistency with standard color temperature~\cite{afifi2019color}. For segmentation, we use SAM, a state-of-the-art model pre-trained for efficient object masking~\cite{kirillov2023segment}. SAM enables us to isolate each apple slice quickly and accurately by adjusting thresholds to generate precise masks, as shown in Figure \ref{f4_model1} (bottom left). The dimension of the image data after segmentation and transformation is $224\times224\times3$ to fit in ResNet-18.

\emph{Image data processing with ResNet-18}: We employ ResNet-18 to process the image data. ResNet-18 is a convolutional deep neural network with 18 layers. It uses residual connections to preserve feature integrity across layers, allowing the network to capture complex visual patterns~\cite{he2016deep}. In this setup, ResNet-18 processes each masked apple slice image and transforms it into a $1\times512$ dimensional embedding that encodes visual information on structural changes and color variations related to MC. The detailed architecture is shown in Figure \ref{f4_model1} (top left). 

\emph{Tabular data processing with FC}: The tabular data is processed through an FC neural network, as depicted in Figure \ref{f4_model1} (top right). This FC network comprises three layers: an input layer, a hidden layer, and an output layer. This FC also helps to reshape the size of tabular data from dimension $1\times3$ to $1\times512$, to be consistent with the image data. 

\emph{Data concatenation for MC prediction}: The tabular data embedding is adjusted to match the shape of the image embedding, facilitating seamless integration with the image data. After each data type is processed separately, the tabular and image embeddings are concatenated and passed through an additional FC network with dimension $1\times1024$, where we can adjust the tabular-image-ratio to output the best final MC prediction. This multi-modal fusion network preserves the distinct properties of both data types, enabling a more robust MC prediction by leveraging the strengths of each modality.

\subsection{Baseline models}

\begin{figure}[h]
    \centering
    \includegraphics[width=0.8\columnwidth]{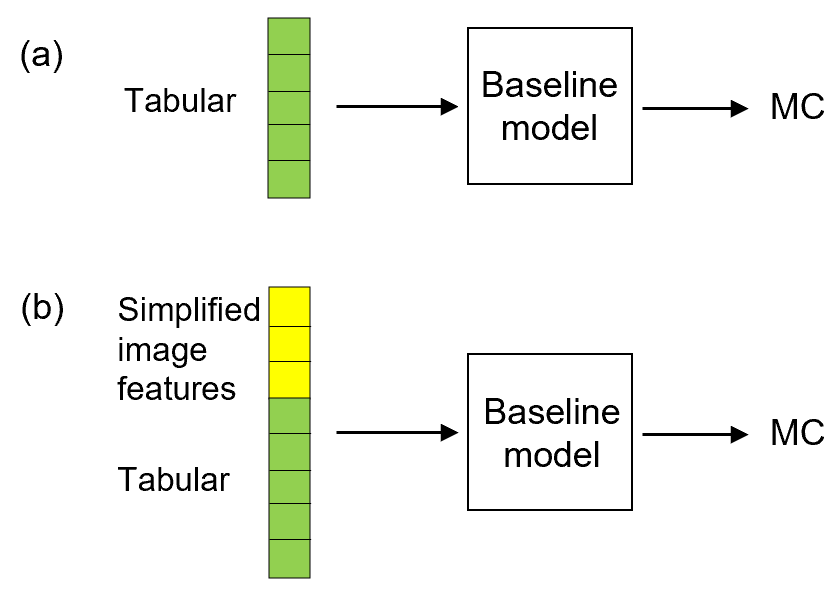}
    \caption{Baseline model architectures for (a) tabular-only data; and (b) standard tabular-image fusion.}
    \label{f5_model2}
\end{figure}

To evaluate the effectiveness of the multi-modal fusion network, we also train baseline model architectures to establish performance benchmarks. The model architectures are shown in Figure~\ref{f5_model2}. Specifically, the first baseline architecture, as shown in Figure~\ref{f5_model2}(a), only uses tabular data for MC prediction, which is commonly referred to as response surface modeling in the literature, with the input dimension at $1\times3$. The second baseline architecture uses a standard data fusion strategy, where simplified image features (average RGB, pixel-wise area) are fused with tabular data as supplementary inputs, with the input dimension at $1\times5$. We implement three commonly used models for each baseline architecture. Linear regression models the relationship between input-output using a linear equation~\cite{montgomery2021introduction}. Gaussian Process (GP) is a non-parametric, probabilistic model that defines a distribution over functions and provides uncertainty estimates by modeling relationships between data points using a covariance function~\cite{hu2023transfer}. NN uses an FC structure similar to the ones used in the multi-modal fusion framework.

\subsection{Ablation study}

\begin{figure}[h]
    \centering
    \includegraphics[width=1.0\columnwidth]{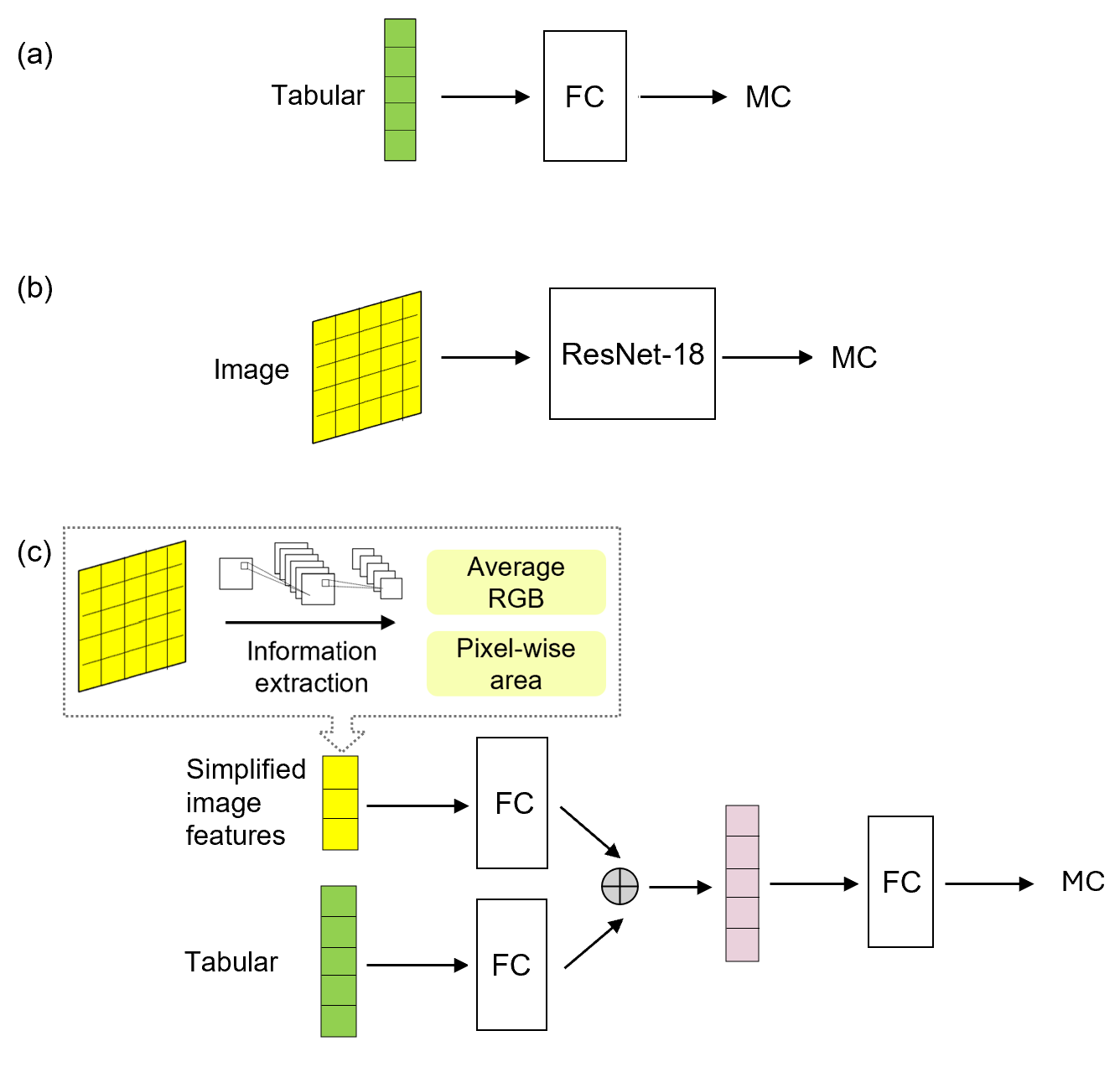}
    \caption{Ablation study for (a) tabular-only data; (b) image-only data; and (c) tabular data with simplified image features.}
    \label{f6_model3}
\end{figure}

An ablation study is conducted to investigate the contribution of each data modality and model component. We investigate the impact of different data configurations on the prediction of MC by comparing three model variations, which are shown in Figure~\ref{f6_model3}.

\emph{Tabular-only}: In this setup (Figure~\ref{f6_model3} (a)), we process only tabular data through the FC layers to predict MC.

\emph{Image-only}: In this setup, which is shown in Figure~\ref{f6_model3}(b), ResNet-18 is used to process the masked image data independently to predict MC to assess the prediction effects of image data alone without using any tabular input.

\emph{Tabular data with simplified image features}: In this configuration, we fuse tabular data with simplified image features, which include average RGB values and pixel-wise area extracted from the masked images, with the detailed framework shown in Figure~\ref{f6_model3}(c). Specifically, the original dimension for tabular and image features are $1\times3$ and $1\times2$, each of them passes through a FC and being matched to the same dimension before concatenation. Different from the standard tabular-image fusion model in the baseline, here, the simplified image features are processed in parallel with the tabular data and concatenated through an FC layer to predict MC, therefore permitting flexible adjustment of the tabular-image-ratio. Although some high-dimensional details may be lost due to dimensionality reduction, this configuration allows us to evaluate the added predictive power of basic image information via simplified multi-modal data fusion methods.


\section{Results}
This section presents the results for predicting MC of apple slices during the drying process using the proposed framework and baseline methods. Section 4.1 describes the six-fold cross validation methods for data splitting and the evaluation metrics; Section 4.2 reports the performance of the baseline models; and Section 4.3 evaluates the contributions of different data modalities through an ablation study.

\subsection{Data split and evaluation metric}

\begin{figure}[h]
    \centering
    \includegraphics[width=1.0\columnwidth]{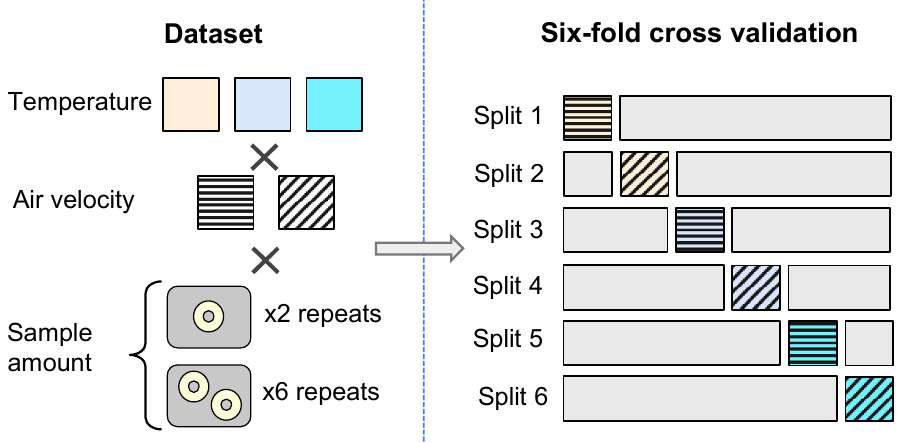}
    \caption{Procedure of data split and six-fold cross validation.}
    \label{f7_result1}
\end{figure}

A six-fold cross-validation approach is used to rigorously evaluate all models. As described in Section 2.1, our dataset includes three levels of temperature and two levels of air velocity, resulting in six distinct combinations of drying conditions. Each fold uses one combination as the evaluation set, with the remaining five combinations as the training set, as illustrated in Figure~\ref{f7_result1}. This split avoids any overlap of similar conditions in the training and evaluation sets, thus ensuring a rigorous assessment of model performance. We use the average RMSE across all six folds as the primary evaluation metric.

\subsection{Baseline performance comparison}

Table~\ref{t2} compares the average RMSEs of the baseline models and the proposed data fusion framework. Among these baseline models, NN achieves the best performance, with an RMSE of 0.0450 when using only tabular data and a lower RMSE of 0.0428 when additional image features are included. For both linear regression and GP, the addition of extracted image features slightly increases the RMSE. This may be caused by the limited capacity of these simpler models to leverage additional image-derived information effectively. These results demonstrate that while basic image features can contribute to MC prediction, they are best utilized in some specific models such as the NN, which can cope with the increased data complexity. In comparison, our proposed method achieves the lowest RMSE of 0.0363. The ground truth MC values range between 0.1 and 0.2. Therefore, this low RMSE level indicates an excellent agreement between prediction and ground truth. Furthermore, we note that in our data-splitting strategy, the training and test datasets were generated with different drying parameters, so it is likely that they have different data distributions. This testing strategy is crucial for addressing the challenges posed by non-identically distributed industrial data, making our evaluation both more challenging and realistic. Our method represents a significant RMSE reduction ranging from 15.2\% to 27.4\% over the baseline methods. This substantial improvement highlights the effectiveness of our data fusion approach in leveraging both tabular and image data for more accurate MC predictions.

\begin{table}[h]
\centering
\captionsetup{singlelinecheck=false, aboveskip=1pt}
\caption{Comparison of average RMSE and RMSE reduction between the proposed method and baseline models.}
\vspace{6 pt}
\footnotesize
\begin{tabular*}{\hsize}{@{\extracolsep{\fill}}p{1.8cm}p{2cm}p{1.8cm}p{2cm}@{}}
\toprule
Dataset & Model & Average RMSE & RMSE reduction \\
\midrule
\multirow{3}{*}{Tabular-only} & \begin{tabular}{@{}l@{}}Linear regression\end{tabular} & 0.0464 & 21.8\% \\
                              & GP                & 0.0499 & 27.3\% \\
                              & NN                & 0.0450 & 19.3\% \\
\midrule
\multirow{3}{*}{\begin{tabular}{@{}l@{}}Standard tabular\\ -image fusion\end{tabular}} & \begin{tabular}{@{}l@{}}Linear regression\end{tabular} & 0.0469 & 22.6\% \\
                                                            & GP                & 0.0500 & 27.4\% \\
                                                            & NN                & 0.0428 & 15.2\% \\
\midrule
\begin{tabular}{@{}l@{}}Multi-modal\\data fusion\end{tabular} & Our method & 0.0363 & - \\
\bottomrule
\end{tabular*}
\label{t2}
\end{table}


\subsection{Ablation study results}

To ensure consistency across configurations, we apply the same hyperparameters for each model in the ablation study, as detailed in Table~\ref{t3}. Parameters such as batch size, hidden size, learning rate, and optimizer are kept uniform to isolate the effect of adding image data, ensuring that any observed improvements in performance result from data configuration rather than hyperparameter differences. The final FC layer concatenates tabular and image data in a uniform 8:1 ratio. 

\begin{table}[h]
\centering
\captionsetup{singlelinecheck=false, aboveskip=1pt}
\caption{Hyperparameters for the models in ablation study.}
\vspace{6pt}
\footnotesize
\begin{tabular*}{\hsize}{@{\extracolsep{\fill}}p{5cm}p{3cm}@{}}
\toprule
Hypermeter & Value \\
\midrule
Batch size         & 64           \\
Hidden size        & 1024         \\
Learning rate      & 0.0001       \\
Number of epoch    & 300          \\
Optimizer          & Adam         \\
Metrics            & RMSE         \\
Tabular-image-ratio    & 8:1   \\
\bottomrule
\end{tabular*}
\label{t3}
\end{table}

Table~\ref{t4} presents the results of the ablation study, highlighting the impact of different data modalities on model performance. The model using only tabular data serves as a baseline for assessing the effect of additional data types. When using image-only data, the RMSE is higher than that of the tabular-only model, suggesting that while image data contributes to MC prediction, it lacks the contextual information provided by process parameters. Combining tabular data with simplified image features improves accuracy, validating the effectiveness of data fusion. This configuration achieves a lower RMSE than the standard tabular-image fusion baseline, demonstrating the benefit of adjusting the proportion of different data modalities. Finally, our multi-modal fusion method, which integrates both tabular and high-dimensional image data, achieves the lowest RMSE, underscoring its capability to capture both general process effects and sample-specific characteristics. This result highlights the advantage of our approach in achieving more accurate MC predictions.

\begin{table}[h]
\centering
\captionsetup{singlelinecheck=false, aboveskip=1pt}
\caption{Average RMSE and RMSE reduction between the proposed method and other datasets for ablation study.}
\vspace{6pt}
\footnotesize
\begin{tabular*}{\hsize}{@{\extracolsep{\fill}}p{4cm}p{3cm}p{3cm}@{}}
\toprule
Dataset & Average RMSE & RMSE\\
\midrule
Tabular                           & 0.0450 & 19.3\% \\
Image only                        & 0.0479 & 24.2\% \\
\begin{tabular}{@{}l@{}}Tabular data with\\simplified image features\end{tabular} & 0.0409 & 11.2\% \\
\begin{tabular}{@{}l@{}}Multi-modal data fusion\end{tabular}
             & 0.0363 & - \\
\bottomrule
\end{tabular*}
\label{t4}
\end{table}

\section{Discussions}

This section provides an in-depth analysis of the ablation study results, highlighting the advantages of the multi-modal fusion model in predicting MC. Section 5.1 discusses performance evaluation with visual and statistical error comparisons. Section 5.2 explores the model's ability to capture small-scale variability. Section 5.3 examines the stability of model performance with varying proportions of tabular and image data.

\subsection{Performance evaluation and error analysis}

\begin{figure*}[h]
    \centering
    \includegraphics[width=1.4\columnwidth]{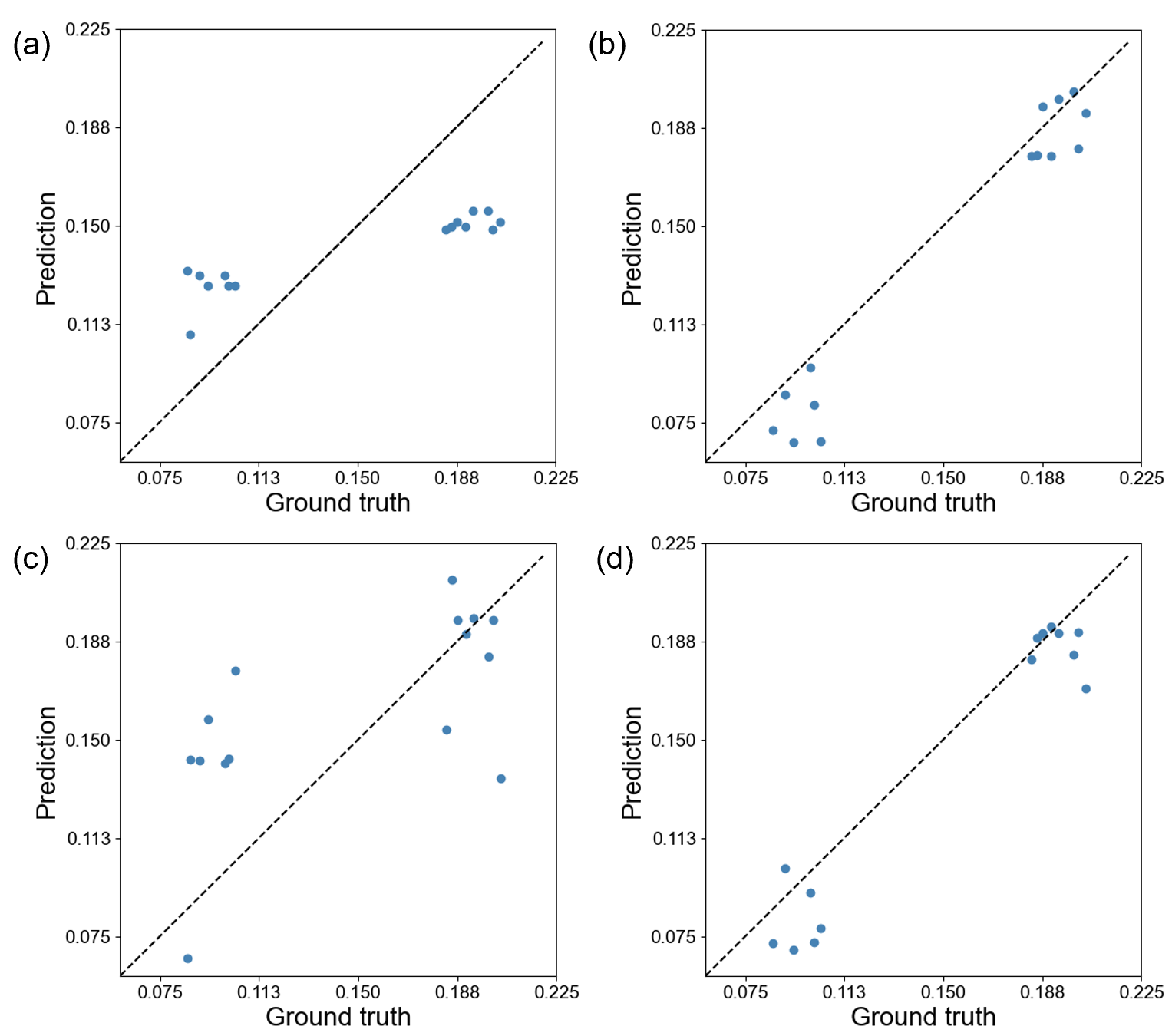}
    \caption{Prediction vs. ground truth for data in 60 $^{\circ}$C, 1.5 m/s for (a) tabular-only, RMSE = 0.0384; (b) tabular data with simplified image features, RMSE = 0.0193; (c) image-only, RMSE = 0.0433; and (d) multi-modal data fusion, RMSE = 0.0178.}
    \label{f8_discussion1}
\end{figure*}

Figure~\ref{f8_discussion1} provides a visual comparison of predicted vs. ground truth MC values across the four configurations in the ablation study, using the evaluation set at 60$^{\circ}$C and 1.5 m/s from the six-fold cross validation. In general, all predictions cluster around the 10\% and 20\% MC levels, showing reasonable alignment with the ground truth distribution.


Comparing Figure~\ref{f8_discussion1}(a) and Figure~\ref{f8_discussion1}(b) shows that fusing simplified image features enhances alignment with the ground truth but introduces greater variability, likely due to the limited details conveyed by these features. Figure~\ref{f8_discussion1}(c) displays greater variance and significant prediction errors for the image-only model, indicating that the image data alone lacks the necessary contextual information for good predictions. Figure~\ref{f8_discussion1}(b) and Figure~\ref{f8_discussion1}(d) exhibit better prediction performance than the other two. These results indicate that multi-modal data fusion substantially enhances prediction accuracy.

\begin{figure}[h]
    \centering
    \includegraphics[width=0.8\columnwidth]{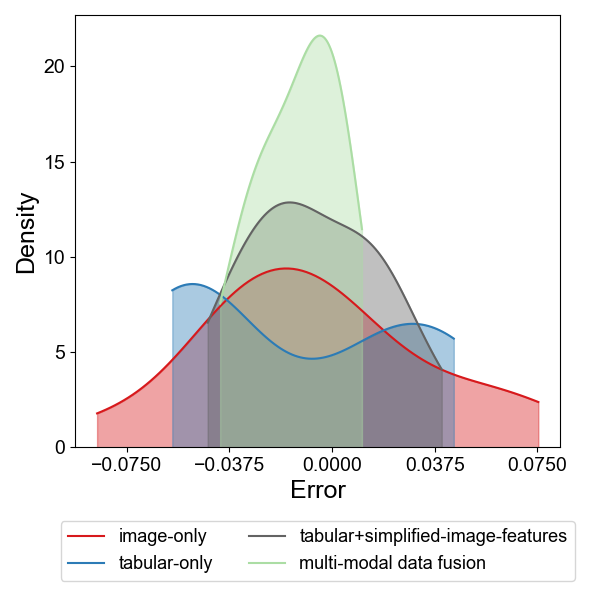}
    \caption{Error density distribution for four models in ablation study.}
    \label{f9_discussion3}
\end{figure}

Figure~\ref{f9_discussion3} further examines the modeling robustness by plotting the density distributions of prediction errors, which are calculated by prediction minus ground truth, for each configuration. The image-only model exhibits the widest error range, indicating instability when image data is used without complementary tabular information. The tabular-only model displays the second-widest range, with errors more dispersed around the ground truth, suggesting that tabular data alone lacks key information present in image data. In comparison, both the tabular-with-simplified-image and multi-modal data using high-dimensional image configurations show reduced error variability, with the latter achieving the narrowest error distribution. These findings reconfirm the superiority of the multi-modal data fusion approach, where effective fusion of two data modalities with preserving high-dimensional information from distinct data types can significant enhance the prediction accuracy and robustness.

\subsection{Capturing small-scale variability}

Small-scale variability refers to subtle differences in drying characteristics that arise from natural variations among samples~\cite{zambrano2019assessment}. These slight variations, even when samples are subjected to identical drying conditions, can affect the final MC values. Capturing this variability is essential for accurately predicting MC, especially in real-world applications where minor differences among samples can affect overall product quality and consistency~\cite{banerjee2024improving}.

\begin{figure}[h]
    \centering
    \includegraphics[width=0.8\columnwidth]{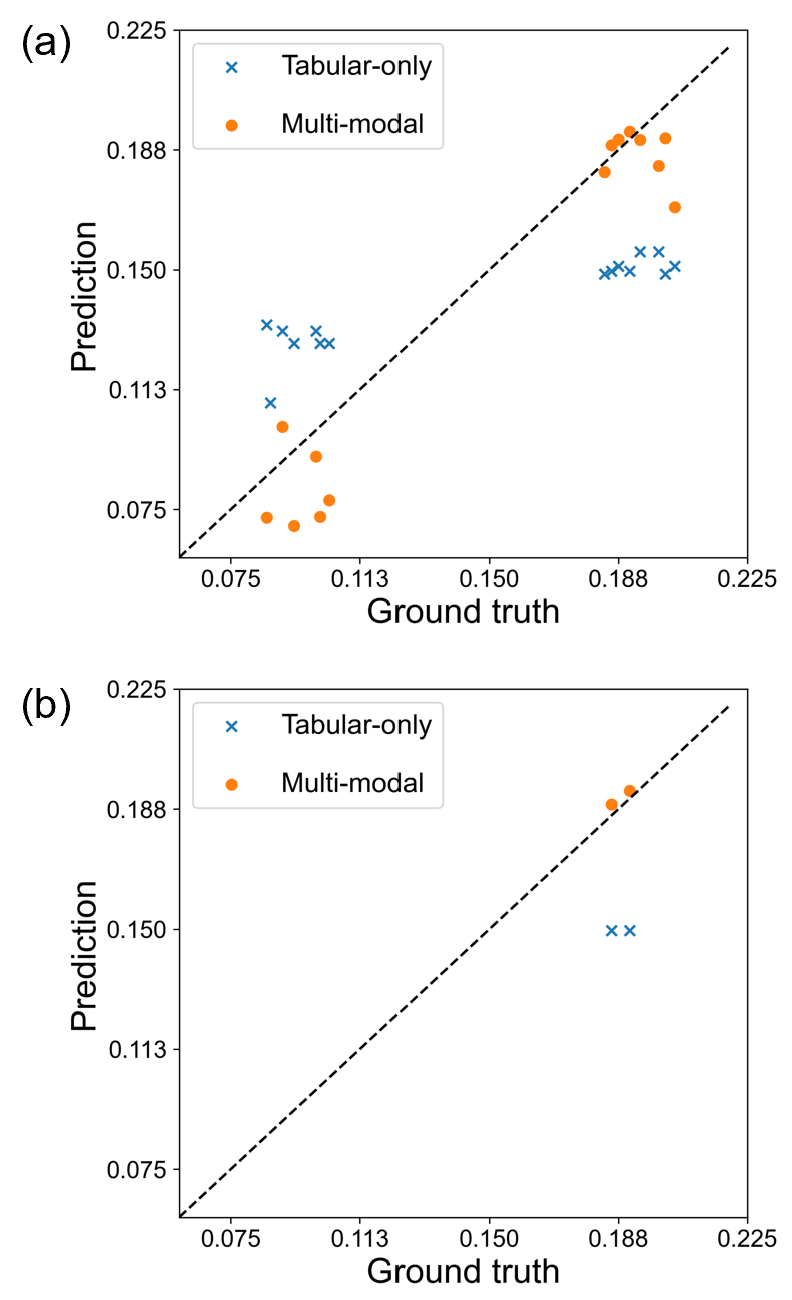}
    \caption{Examining small-scale variability with a comparison of tabular-only data and multi-modal data under drying conditions at (a) (60$^{\circ}$C, 1.5m/s) and (b) (60$^{\circ}$C, 1.5m/s, 140mins).}
    \label{f10_discussion3}
\end{figure}

To evaluate the model's ability to capture small-scale variability, we use an evaluation set at 60$^{\circ}$C and 1.5 m/s, focusing on two apple slices dried together under the same conditions (i.e., temperature, air velocity, and drying time), as an illustrative example. Figure~\ref{f10_discussion3} compares the predictions vs. ground truth MC values for the tabular-only model and the multi-modal fusion model. In Figure~\ref{f10_discussion3}(a), we notice that the predictions using only tabular data are identical for both apple slices dried together, as tabular data alone cannot distinguish between individual sample characteristics. In contrast, predictions using both tabular and masked image data vary between the two slices, aligning more closely with the ground truth. For example, Figure~\ref{f10_discussion3}(b) demonstrates that under drying conditions at 60$^{\circ}$C, 1.5m/s, 140 mins the tabular-only model predicts the same MC for both slices, showing deviation from the actual values. Meanwhile, the multi-modal fusion model accurately differentiates between the two slices, with predictions closely matching the real MC values.

The multi-modal fusion model demonstrates enhanced capability in accounting for sample-specific differences. This advantage is particularly valuable in applications where consistency and precision are essential, underscoring the effectiveness of multi-modal fusion for handling complex patterns in MC prediction.

\subsection{Impacts of the tabular and image portions in data fusion}

Our multi-modal data fusion method allows for an adjustable ratio between each distinct data modality, making the tabular-to-image ratio a unique hyperparameter in our fusion framework. This flexibility prompts an examination of how varying this ratio impacts prediction accuracy. Specifically, we have configured the framework so that a 1:1 tabular-to-image ratio represents equal dimensionality for both data types.

\begin{figure}[h]
    \centering
    \includegraphics[width=0.9\columnwidth]{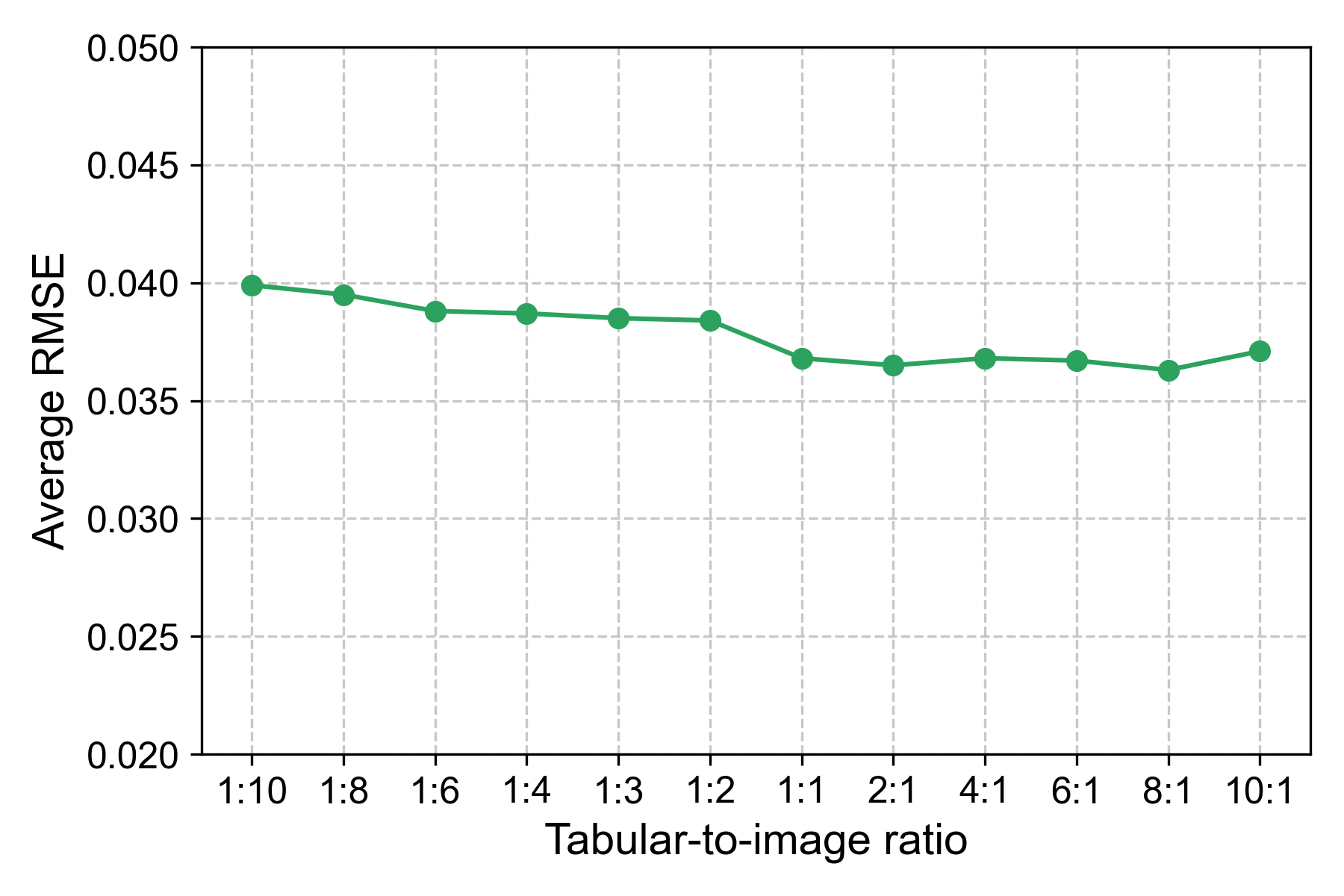}
    \caption{Average RMSE trend under tabular-image-ratios from 1:10 to 10:1.}
    \label{f11_discussion4}
\end{figure}

To assess the model's stability and robustness, we adjust the proportion of tabular data relative to high-dimensional image data in the final fusion layer, testing extreme cases of 1:100 and 100:1, as well as intermediate ratios from 1:10 to 10:1. The results in Table~\ref{t5} show that the average RMSE remains relatively stable across ratios from 1:10 to 10:1, with optimal performance observed at an 8:1 tabular-to-image ratio. These fluctuations are visualized in Figure~\ref{f11_discussion4}. In general, average RMSE is lower when the proportion of tabular data is equal to or exceeds that of high-dimensional image data, suggesting that tabular data plays a more significant role in MC predictions.

At the extremes of 100:1 and 1:100, where the model approaches tabular-only and image-only configurations, we observe increased RMSEs. This finding suggests that while tabular data provides valuable contextual information, it cannot fully replace the rich feature information captured by image data, and vice versa. This analysis underscores the importance of multi-modal data fusion in enhancing MC prediction accuracy and demonstrates that an adjustable ratio among data modalities can improve model performance. The model’s stability across a range of ratios highlights its ability to balance both data types effectively for accurate predictions, even as the fusion ratio varies.

\begin{table*}[h]
\centering
\captionsetup{singlelinecheck=false, aboveskip=3pt}
\caption{Average RMSE for different tabular-to-image ratios.}
\footnotesize
\begin{tabular*}{\textwidth}{@{\extracolsep{\fill}}lllllllll@{}}
\toprule
Tabular-to-image ratio & 1:100 & 1:10 & 1:8 & 1:6 & 1:4 & 1:3 & 1:2 & 1:1 \\
\midrule
Average RMSE         & 0.0440 & 0.0399 & 0.0395 & 0.0388 & 0.0387 & 0.0385 & 0.0384 & 0.0368 \\
\cmidrule[1pt](lr){1-9} 
Tabular-to-image ratio & 2:1 & 4:1 & 6:1 & 8:1 & 10:1 & 100:1 \\
\midrule
Average RMSE           & 0.0365 & 0.0368 & 0.0366 & 0.0363 & 0.0371 & 0.0400 \\
\bottomrule
\end{tabular*}
\label{t5}
\end{table*}

\section{Conclusion and future work}

This study presents a multi-modal data fusion framework for predicting the final MC of apple drying. By combining tabular data from drying conditions with high-dimensional image data that captures sample-specific features, the model utilizes parallel processing through FC layers and ResNet-18 to retain contributions of each data source and mitigate information loss. Experimental studies reveal that the multi-modal fusion model achieves notable RMSE reductions--19.3\%, 24.2\%, and 15.2\% over tabular-only, image-only, and standard tabular-image fusion models, respectively. The model also shows the flexibility of adjusting portions of each data modality for better predictions, and demonstrates stability across varying tabular-to-image data ratios, effectively capturing sample-specific variability, which underscores its adaptability to complex data configurations. These findings highlight the effectiveness of multi-modal fusion for improving predictive accuracy in drying processes, where precise MC prediction is essential for quality control and efficiency improvement. 

There are several promising directions for future research. From an application perspective, incorporating the effects of apple varieties could enhance the generalizability of the proposed approach, as the current study focuses solely on Fuji apples. To capture the effect of apple types, we may simply add a categorical variable or characterize apples with physical, chemical, and textural properties. The multi-modal data fusion framework developed in this paper is readily extensible to incorporate these additional data modalities. Furthermore, optimizing the design of apple drying equipment based on insights from the proposed model could lead to significant industrial cost savings. From a methodological standpoint, we may extend this multi-modal fusion approach with its encoder-decoder architecture to other drying technologies and products, and investigate its scalability across varied manufacturing applications. Additionally, incorporating this approach for online monitoring of drying process could further broaden the model’s applicability.

\section*{Acknowledgements}

This study was supported by the Center for Advanced Research in Drying (CARD). 






\bibliography{reference}
\bibliographystyle{elsarticle-num}












\clearpage\onecolumn

\normalMode






\end{document}